\pdfoutput=1

\documentclass[pdflatex,sn-basic]{sn-jnl}

\usepackage{times}
\usepackage{epsfig}
\usepackage{graphicx}
\usepackage{amsmath}
\usepackage{amssymb}
\usepackage{algpseudocode}
\usepackage{algorithm}
\algnewcommand\algorithmicforeach{\textbf{for each}}
\algdef{S}[FOR]{ForEach}[1]{\algorithmicforeach\ #1\ \algorithmicdo}

\usepackage{cellspace}
\usepackage{amsfonts}
\usepackage{physics} 
\usepackage[protrusion=true,tracking=true,kerning=true,expansion,final]{microtype}   
\usepackage{booktabs} 
\usepackage{subcaption}
\usepackage{nicefrac} 
\usepackage{cancel}

\usepackage{xspace}
\usepackage{upref}
\usepackage{textcomp}
\usepackage[T1]{fontenc}

\usepackage{array}
\usepackage{amssymb}
\usepackage{gensymb}
\usepackage{bbding}
\usepackage{mathtools}
\usepackage{centernot} 
\usepackage{bigints}
\usepackage{dsfont} 
\usepackage{esint} 
\usepackage{multirow}
\usepackage{enumitem}
\usepackage{graphbox} 
\usepackage{xcolor}

\usepackage{algorithm}
\usepackage{algpseudocode}
\usepackage{algorithmicx}

\usepackage{varioref} 

\usepackage[capitalise]{cleveref} 
\crefname{appsec}{appendix}{appendices}
\Crefname{appsec}{Appendix}{Appendices}

\usepackage{url}

\renewcommand{\cite}[1]{\citep{#1}}
\definecolor{mydarkblue}{rgb}{0,0.08,0.45}
\definecolor{urlcolor}{rgb}{0,.145,.698}
\definecolor{linkcolor}{rgb}{.71,0.21,0.01}
\hypersetup{ %
	pdftitle={GoToNet: Fast Monocular Scene Exposure and Exploration},
	pdfauthor={},
	pdfsubject={}, 
	pdfkeywords={},
	pdfborder=0 0 0,   
  pdfpagemode=UseOutlines,
	breaklinks=true, 
	colorlinks=true,
	bookmarks=true,
	bookmarksnumbered=true,
	urlcolor=urlcolor,
	linkcolor=linkcolor,
	citecolor=mydarkblue,
	filecolor=mydarkblue,
	pdfview=FitH}

\vbadness=\maxdimen
\usepackage{bm}
\newcommand{\eb}[1]{{\scriptsize\,$\pm$\,#1}}

\jyear{2021}%

\theoremstyle{thmstyleone}%
\theoremstyle{thmstyletwo}%

\theoremstyle{thmstylethree}%

\begin{document}

\title[GoToNet: Fast Monocular Scene Exposure and Exploration]{GoToNet: Fast Monocular Scene Exposure and Exploration}


\author*[1]{\fnm{Tom} \sur{Avrech}}\email{tomavrech@gmail.com}

\author[1]{\fnm{Evgenii} \sur{Zheltonozhskii}}\email{evgeniizh@campus.technion.ac.il}

\author[1]{\fnm{Chaim} \sur{Baskin}}\email{chaimbaskin@cs.technion.ac.il}

\author[1]{\fnm{Ehud} \sur{Rivlin}}\email{ehudr@cs.technion.ac.il}

\affil[1]{\orgdiv{Department of Computer Science}, \orgname{Technion -- Israel Institute of Technology}, \orgaddress{\city{Haifa}, \postcode{3200003}, \country{Israel}}}

\abstract{Autonomous scene exposure and exploration, especially in localization or communication-denied areas, useful for finding targets in unknown scenes, remains a challenging problem in computer navigation.
In this work, we present a novel method for real-time environment exploration, whose only requirements are a visually similar dataset for pre-training, enough lighting in the scene, and an on-board forward-looking RGB camera for environmental sensing. As opposed to existing methods, our method requires only one look (image) to make a good tactical decision, and therefore works at a non-growing, constant time.
Two direction predictions, characterized by pixels dubbed the Goto and Lookat pixels, comprise the core of our method. These pixels encode the recommended flight instructions in the following way: the Goto pixel defines the direction in which the agent should move by one distance unit, and the Lookat pixel defines the direction in which the camera should be pointing at in the next step. These flying-instruction pixels are optimized to expose the largest amount of currently unexplored areas.

Our method presents a novel deep learning-based navigation approach that is able to solve this problem and demonstrate its ability in an even more complicated setup, i.e., when computational power is limited.
In addition, we propose a way to generate a navigation-oriented dataset, enabling efficient training of our method using RGB and depth images.
Tests conducted in a simulator evaluating both the sparse pixels' coordinations inferring process, and 2D and 3D test flights aimed to unveil areas and decrease distances to targets achieve promising results.
Comparison against a state-of-the-art algorithm shows our method is able to overperform it, that while measuring the new voxels per camera pose, minimum distance to target, percentage of surface voxels seen, and compute time metrics.}




\maketitle
\section{Introduction}

\begin{figure}[htp]
\centering
\includegraphics[width=\linewidth,keepaspectratio]{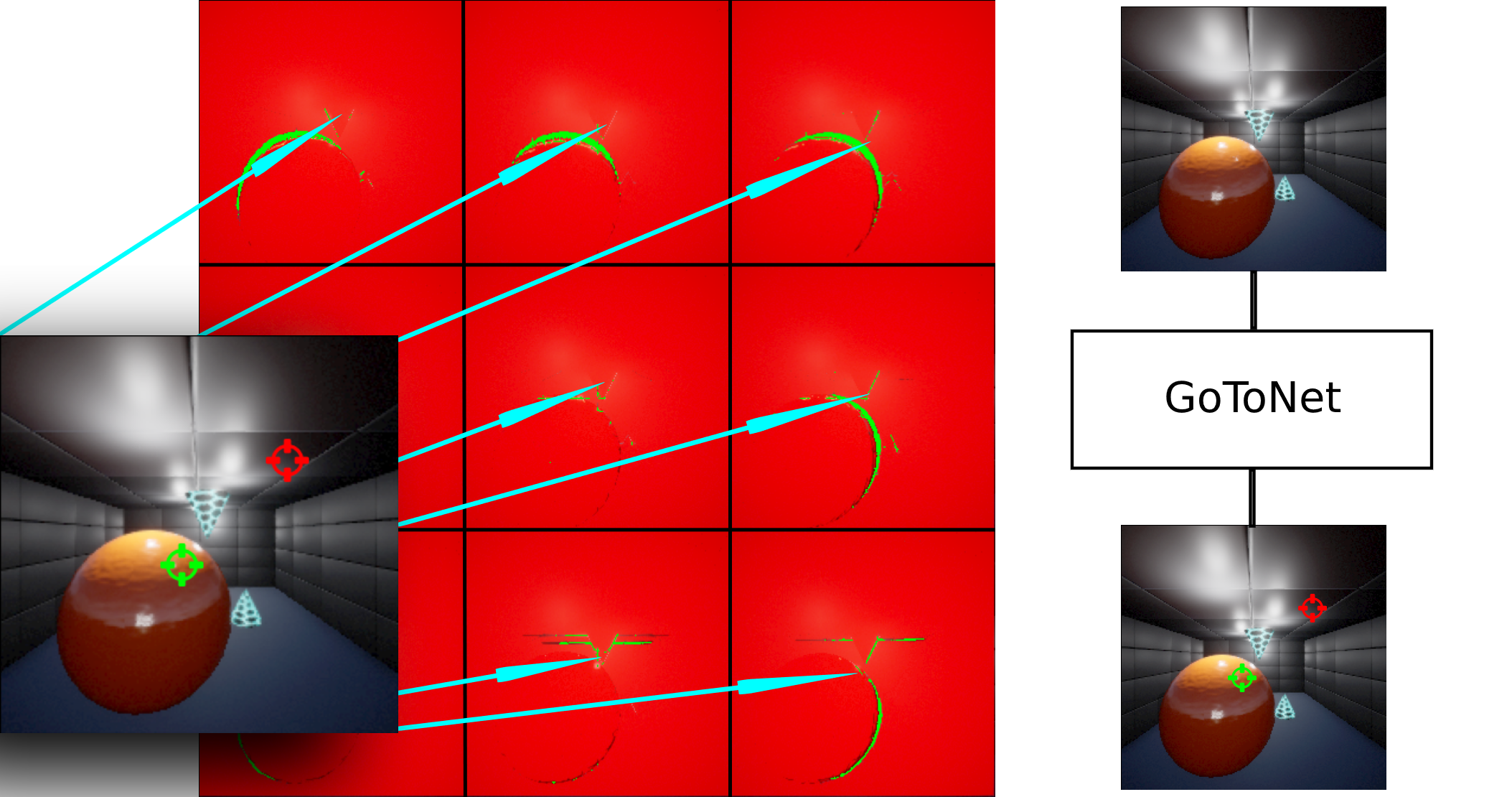}
\caption{(Better viewed in color) 
\textbf{Left}: Data creation. RGB-D capture is made from a 3D pose. Using ray tracing, the currently visible areas are painted red. Then $3\times 3$ 3D camera poses are calculated, representing $3\times 3$ different movement directions. An RGB image is captured for each pose. Pixels that represent unseen areas are colored green, counted, and used to calculate both the Goto (red crosshairs), and Lookat (green crosshairs) pixels. 
\textbf{Right:} Navigation model gets a RGB image as an input and predicts both the Goto and Lookat pixels.}
\label{fig:data_creation}
\end{figure}

Modern monocular-based simultaneous localization and mapping (SLAM) methods \cite{padhy2019localization, Buyval, Celik, 8452299}, such as Bundle Adjustment (BA) \cite{10.1007/3-540-44480-7_21, chen2019bundle}, allow navigation in environments with limited localization tools (e.g., GPS-denied indoor environments), or previously unknown environments. These methods, however, have high power and runtime requirements, making them unsuitable for devices with limited computational power, such as Micro Aerial Vehicles (MAV) and smartphones, and for real-time applications.
As a result, many existing indoor navigation methods rely on the assumption that the navigator can use sensors that already exist in that environment (e.g., visual markers \cite{Michel}, Wi-Fi routers’ positions \cite{Chang}, Bluetooth beacons’ positions \cite{Mazan2015ASO}, radio-frequency identification (RFID) tags \cite{Reza}, magnetic positionings \cite{HAVERINEN20091028} or even have maps of that environment \cite{Geva}. Another option to improve this demanding process is to support the SLAM algorithm by making the exploration process more efficient \cite{indelman2012incremental}.

One possible solution for exposing and exploring a scene is the use of a swarm of MAVs, such that each individual agent may possess smaller amounts of sensors or computational power, yet is considerably cheaper to operate. This allows for a large amount of low-cost agents to work simultaneously and achieve a goal. This allows to have a part of the agents lost without compromising the main objective of the mission. This approach of a swarm of agents enables covering a large area, for example for search-and-rescue missions, which would require the classic approach longer time to complete, or even fail if given a time-critical mission (e.g., rescue an injured individual).

In this work, we present a novel method for real-time environment exploration, whose only requirements are a visually similar dataset for pre-training, enough lighting in the scene, and an onboard forward-looking RGB camera for environmental sensing. As opposed to existing methods, such as BA \cite{10.1007/3-540-44480-7_21, chen2019bundle}, our method requires only one look (image) to make a good tactical decision, and therefore works at a non-growing, constant time (visualized in Figure \ref{fig:data_creation}).

\begin{figure}[htp]
\centering
\includegraphics[width=\columnwidth,height=2.8cm,keepaspectratio]{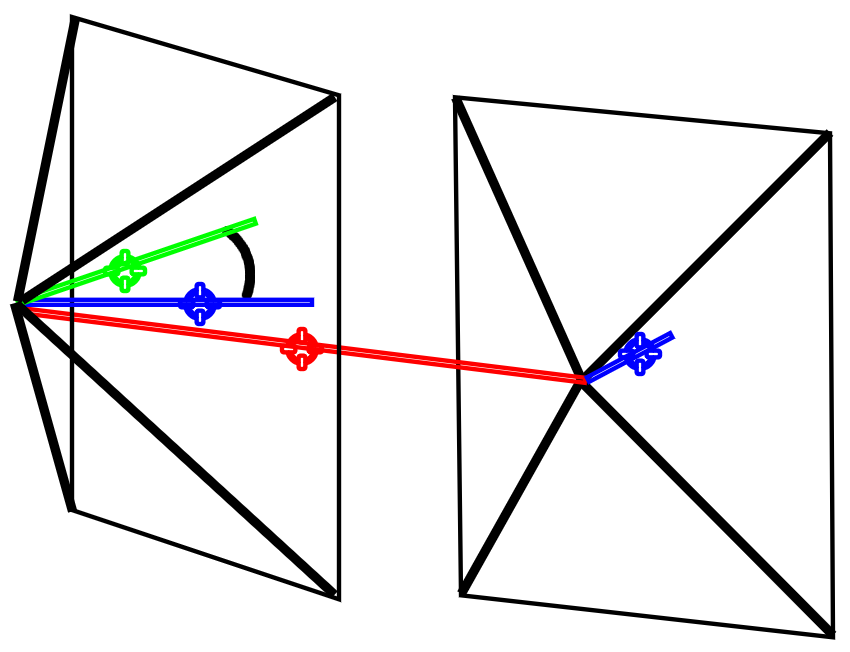}
\caption{An illustration of navigation using Goto and Lookat pixels. The left camera pose represents the initial situation, and the blue vector is the principal axis. The Goto (red) pixel and the Lookat (green) pixel are predicted, and back-projected into vectors. In this case, they translate into the instructions ``go right and look left''.  Then, the camera moves to the right camera pose, whose position is at a fixed distance along the Goto pixel direction, and is rotated by an angle determined by the Lookat pixel direction.}
\label{fig:goto_lookat_movement}
\end{figure}

Two direction predictions, characterized by pixels dubbed the Goto and Lookat pixels, comprise the core of our method. These pixels encode the recommended flight instructions in the following way, (visualized in Figure \ref{fig:goto_lookat_movement}): the Goto pixel defines the direction in which the agent should move by one distance unit, and the Lookat pixel defines the direction in which the camera should be pointing at in the next step (taking into account movement dictated by the Goto pixel).

The Goto and Lookat pixels inference method tries to select those 2 pixels in a way that optimizes exposure. What should be our flight's next pose such that a picture taken from that next step displays the maximum amount of pixels that represent areas unseen from our current location ? We wish we could check all possible next poses, and choose the pose that maximizes exposure. This is too computational expensive. For now, we evaluate only a very sparse set of such poses, and weight-average them.

In order to research our approach, a supporting environment was built. This includes: For the scene painting, a framework was built on top of the AirSim framework \cite{shah2017airsim}, which adds polygons into the scene for rendering. For the pixels' coordinations inferring, which also evaluates it's performances against a desner case, a camera poses placer and a rendering framework was created. For the deep neural networks' training, we built a grid-search algorithm, which searches for the closer-to-optimal hyper-parameters for our networks. For training, we also built a custom data augmentation method. For the test flights, we built a framework which places the drone in the scene, takes our pretrained model, infers commands, and executes them, according to the limitations described in the experiments section.

Our main contributions are:
\begin{itemize}
    \item A new approach for scene exploration using RGB input by inferring movement and camera directions. These directions are encoded in Goto and Lookat pixels.
    \item A novel Microsoft AirSim (Unreal Engine) based \cite{shah2017airsim} dataset generator, with samples consisting of a RGB image, corresponding depth image and Goto and Lookat pixel coordinates optimized for a high scene exposure flight. 
    \item An autopilot-like framework that predicts Goto and Lookat pixels from RGB images.
    \item A deep neural network (DNN) trained to predict the pixels on our RGB-D dataset in a supervised way, which does not require a depth sensor for inference (but can utilize such sensor, if available).
\end{itemize}

In the following sections, we give some background (\cref{sec:background}) for the exploration and exposure problems that we try to solve in this work. In addition we describe related works (\cref{sec:related}) that do similar attempts to solve those problems. We than describes our approach (\cref{sec:method}), and how we implement it. Then we describe tests done in order to evaluate and compare our approach (\cref{sec:experiments}). Finally, we concludes our work and propose several directions for possible future works (\cref{sec:conclusion}).

\section{Background}
\label{sec:background}
Autonomous navigation is different from the traditional human-controlled approach in many senses, since a human can understand the scene significantly quicker and better, and, therefore, better control the vehicle. As in any other field, the constant trend is to try and replace humans by computers so they can be doing the same jobs, as good as or even better than humans in all metrics. Therefore, significantly decrease the need for workers and both the quality and quantity of errors. In our case - be able to navigate in the same or better quality without a real-time pilot.

The common approach used for unknown scene mapping using a monocular camera is the Bundle Adjustment method \cite{10.1007/3-540-44480-7_21}. This method receives an amount of RGB images as it's input, and builds a 3D map of the environment, including the camera's current and past 3D poses. It does it in the following way: Using some detection algorithm, it detects 2D features laying upon all of the images. A feature is made of a pixel's coordinations, and some sequence of bits that identify it. Features matching is executed between couples of images. This is done in order to know that a feature is seen in both given images. Using the list of matching features, and some initialization algorithm, a 3D map of both the 3D features and the 3D camera poses is created. Now the main BA algorithm takes place. This algorithm solves a minimization problem. It minimizes the average Euclidean distance between 2D features detected previously upon all images, and their correspond projected pixels, which were projected using the 3D map. At each iteration of this algorithm, the 3D map is updated, both by updating the 3D locations of the features, and the camera's 3D poses. This is done until convergence. The output is the 3D map estimating both the features' and the camera' poses. A known issue with this approach, is that it doesn't provide a reliable scale estimation. This means we can have a good overall estimation about the relations between the features' and the camera's locations, but an unreliable estimation about the distances in between all of them. To try to reduce the scaling issue, more approaches were developed, described in the related work section. These approaches use extra inputs to their systems, such as the usage of more sensors (e.g. a depth camera), or they relay on having previous knowledge about the structure of the specific scene (e.g. a floor plan). Another issue with this approach is it's computational demands. For example, as more images stack up, more features are required to be compared to all previous detected features. This prevents the usage on many real-time systems.
This algorithm doesn't demand the usage of sequential data (video), or even the usage of the same camera at any input image. For navigation, we will probably have the simpler case of having the same camera capturing a sequence of images, and therefore having many matching features in between every 2 sequential images.
Our method does not require a classical approach like this, which creates a map of the scene.

Methods for navigation in unknown scenes try to maximize knowledge about the scenes' structures or find targets in them. As an agent suffers from uncertainly, and may be limited in many ways, such as relaying only on it's on-board sensors, battery time, mission expiration time, or holding not accurate/up-to-date knowledge about the structure of the scene, it has to choose it's next steps in a well-thought-of way. The term Next-Best-View describes this. What should be the agent's next step, out of all possible next steps, such that hopefully it will get the agent closer to it's goal, as much as possible. In our work, the agent is limited by only having a forward looking RGB camera, previous knowledge about how the scene may look like, and what should be the actions accordingly.

Deep learning is a term that describes a vest amount of deep neural networks (DNN). In this work, we focus on DNNs that get images as their inputs. The architecture of these networks contains a large amount of 2D filters. These networks are made of layers upon layers, such that each layer is the result of mathematical operations between it's previous layer and the layer's filters. Those filters detect specific features, and as we look deeper into the network, the filters detect richer features. For example, at the start of the DNN the filters may detect simple lines, and at the end of the DNN the filters may detect a wheel of a car. Unlike the pre deep learning era, in order to make a DNN work well, training using a dataset is required. That means, for example, that if we want to be able to detect whether a picture shows a cat or a dog, we will need to have a dataset with many examples of how both cats and dogs look like. Therefore, the result of a DNN's prediction takes into account all the data contained in the input. Like anything else in computers, both the inputs (e.g. images) and the outputs (e.g. label) are made of binary numbers. This allows for the inputs and outputs to be anything. A main contribution of our work is taking the VGG16 network \cite{simonyan2014very}, and modify it such that as an input it receives a gradients map, and it outputs 2 pixels' coordinations. Another DNN used in our work is the DenseDepth network \cite{alhashim2019high}, which receives a RGB image as an input, and outputs a depth image.
\section{Related Work}
\label{sec:related}
Since our approach of predicting flight instructions via pixels predictions to expose hidden areas is novel, we searched in a broad spectrum of similar methods. In this section, we describe works from the scene mapping and reconstruction, navigation in unknown scenes, collision avoidance and object edges detection, and deep learning for feature extraction and depth prediction fields.

\subsection{Scene mapping and reconstruction}
Mapping or reconstructing a scene is often necessary for performing tasks in unknown scene, e.g., finding an object in the scene. However mapping usually requires large amount of resources and computation. One popular method for scene mapping is Bundle Adjustment \cite{10.1007/3-540-44480-7_21}. Bundle adjustment is the problem of refining a visual reconstruction to produce jointly optimal 3D structure and viewing parameter. This problem is not new, but keeps getting better solutions, such as \cite{Demmel_2021_ICCV}, which better suites for real-time application. Additional approaches to scene mapping include predicting volumetric occupancy with corresponding semantic labels from a single-view RGB-D images \cite{dourado2020edgenet}, or like \cite{DBLP:journals/corr/abs-2011-03981} which uses 3D voxels maps, and predicts the maps' unknown parts for maps completion/reconstruction.

\citet{9197388} presented a LiDAR-based large-scale volumetric urban scene reconstruction method, taking loop-closures in consideration, which creates a high-detailed 3D meshed representation of scenes. Testing show the method works at the scales of at least single digit Kilometers.

\citet{DBLP:journals/corr/abs-2105-11344} uses a DNN for loop closure detection, using data generated using a LiDAR sensor. Their system's input is made of data concluded from 2 LiDAR scans. The data concluded includes 1-channel depth maps, 3-channel normal maps, 1-channel intensity/remission map, and 3D semantic classes probabilities tensor. They estimate both how well the 2 scans overlap, and the Yaw angle differences between 2 LiDAR readings.

\citet{10.1007/978-3-030-58621-8_26} combines both classical SLAM (Bundle Adjustment) and RGB to depth predictions to mutual complement each method's weaknesses and result in more refined mapping. They do that using a self-improving strategy, which both updates the depth prediction model, and the pose prediction from the SLAM algorithm. It first takes RGB data, predicts a depth map, incorporate it to refine the camera's pose prediction. Then, it uses the new pose estimation to update the depth prediction model using the SLAM algorithm's depth estimations.

\citet{8894002} Tries to improve SLAM algorithms by tackling the used often assumption the environment contains only static, non-moving objects, and therefore 3D features that lay upon those objects. To do that, they use a pixel-wise segmentation DNN to detect dynamic objects. They don't only find those feature every image as a stand-alone, but also compare them to the next image of the sequence, in order to increase accuracy, and the limitations of the segmentation DNN. Then, they ignore features laying on those objects, so only static features remain for a more robust SLAM result.

Many of the above methods do good jobs while using different sensors. In the long term, we believe a RGB camera provides the largest amount of information, which includes much more than just distances. Therefore, we believe Bundle Adjustment is going to over-live most other approaches, as one of it's biggest issues, computational power, will decrease with futuristic more powerful computers.

\subsection{Navigation in unknown scenes}
Navigation is closely related to scene mapping, and each of those problems can be used to better the solution of the other: better navigation allows faster scene mapping, while navigation in a mapped scene is significantly simpler.

The most relevant paper to our work is Learn-to-Score \cite{hepp2018learntoscore}. Learn-to-Score predicts future viewpoints' quality, with an approach named next-best-view. Next-best-view approach chooses the action (in case of navigation -- movement or rotation) which maximizes some metric, namely, scene exposure. This is done on a known voxel-represented map, using a 3D convolutional neural network. Unfortunately, the usage of voxels severely limits the possibility of usage in real-world scenarios, where acquiring voxel data would be challenging, especially in real-time settings.
For movement, 1 out of 9 possible movements (up, down, left, right, forward, backward, rotate by -25, +25, or 180 degrees) is chosen according to the score that determines the uncertainty about the world. The discreteness of the action space limits the drone's movement. For example, it is hard to move in a direction that is not a multiplication of 25 degrees or to get to an altitude that is not a multiplication of the movement unit.
Finally, this method calls for depth cameras, which are significantly heavier and larger than RGB cameras. This requirement limits the choice of the drone, its battery life, speed, maneuverability, etc.

Multiple similar methods use different kinds of depth sensors.
\citet{sax2019midlevel} uses multiple filters on a RGB image input, such as corners detector, 2D edge detector, etc. to decide which filters maximize the reward for some unknown motor commanding task. They test their approach on number of tasks, such as navigation to a target, local planning, and visual exploration. In the visual exploration, they use range-finding lasers to detect and unlock unexplored cells in the scene.
\citet{wang2020explore,nguyen2020autonomous} uses a depth camera. They use a historic 3D voxels-based reconstruction of the scene, and the current depth capture, translated into a binary - visited/unvisited areas, in order to predict a classification of the next-best-view (up, down, left, or right) the agent should preform, in order to maximize the coverage of unknown areas. 
\citet{MANSOURI2020103472} uses both a LiDAR sensor, and RGB cameras for navigation in underground mines. They use the range-finder to stay at the center of the tunnel, by using vector geometry, such that the left and right looking vectors have as much as possible equal lengths, and the forward looking vector is largest than a minimal length. They also have a CNN, which takes an RGB image, and outputs a left, forward, or right commands. They also relay on an IMU sensor for altitude, attitude and velocities estimations.

\citet{8452299} uses RGB-D camera for navigation planning for unknown scene exploration. They start with the agent's initial pose, and plan a vector of next poses that will maximize the coverage of all areas of the scene. Although it looks interesting, it seems this paper makes a false assumption. They plan to go to poses that they assume are located in free areas, while their sensors don't see those poses before making the decision to go there.

\citet{chaplot2020neural} navigates to a goal image using topological representations for space that effectively leverage semantics and afford approximate geometric reasoning. The robot receives a panoramic image, and predicts possible movement directions and the direction most likely leading to the target. They model the scene in a graph. If the model thinking the agent and the goal are in the same node, it recommends a more specific location to travel to. Otherwise, it recommends a wider range of possibilities to travel to.

\citet{padhy2019localization} tries to keep an UAV in the center of a corridor, using monocular camera's data. They do that while trying to fix the UAV's poses along the central bisector line. They use 2 DNNs. The first DNN is for positional estimations, and the second is for angular estimations. They have also created a dataset for this task.

\citet{Celik} uses a monocular camera for indoor navigation, by using computer vision algorithm to determinate depths, detect vanishing points in corridors, and other 3D features in a scene, and allow a aerial vehicle to fly in them.

\cite{9561902} presents a decentralize multi-UAVs collision-avoidance smooth paths planning in unknown obstacle-rich scenes method, which uses both depth camera and IMU sensors, and is able to work under unreliable communication conditions. They use topological planning to have their paths planned from a wider perspective, and therefore avoid local minima.

\citet{chen2019learning} presents a method for exploring unknown environments using a RGB-D camera and a bump sensor. The framework  builds an estimation map of the scene using the depth data. The actions prediction model is a DNN, which receives a RGB image, and a 2D map of the estimated location of the robot, surrounded by the environment. An internal policy is updated during training by the approximate map and a collision penalty, where the reward depends on exposure and collision. Actions for the agent are then predicted, in order to maximize the coverage of the scene.

Additional approaches that are worth mentioning, use reinforcement learning for navigation \cite{9102361, zhu2016targetdriven,wahid2019long,ma2019using}. 

In this section, we covered approaches that may look for targets, or expose the scenes, using multiple types of sensors, and models. Some of those methods were built for specific, narrow problems, and some for more general cases. We believe RGB cameras based methods have the advantage, as they can both contain semantic information, and can be used for depth estimations, and therefore fit both the specific, and the more general proposes.

\subsection{Collision avoidance and object edges detection}
For the safety of an agent moving inside the scene, it needs to know which areas to avoid. Although the method we present in this work does not explicitly focus on this task, it does work in a manner that tries to avoid obstacles implicitly.

\citet{badki2021binary} is a method which takes 2 consecutive RGB images, and predicts a binary pixel-wise map. A pixel that has a value of True determinates the object that pixel represents is about to collide with the camera's projection plane under a predefined amount of time. This method can work in real-time. The disadvantage here is that we don't know where on the projection plane the object will collide. This means an object can be said to collide, but actually just by passes the camera without colliding, because it's simply too far away from the camera, yet does intersect it's projection plane.

\citet{9102526} uses RGB-D data and the agent's orientation as the input of a Convolutional Neural Network, which then outputs 1 out of 5 possible commands to the agent's controllers, in order to avoid collision with objects.

\citet{wenzel2021visionbased} uses a deep reinforcement learning approach on a simple 2D maze to predict controller actions. They came to the conclusion simple discrete action space provided them with the best results. The use a monocular camera. Data from that camera is transfer into depth maps using DNN.

\citet{9200530} presents a method for an UAV swarm formation-collision co-awareness by adapting the thin-plate splines algorithm to minimize deformation of the swarm’s formation while avoiding obstacles.

\citet{9001167} presents a decentralized multi-UAV collision avoidance under imperfect sensing method, which uses a two-stage reinforcement learning training approach, and tries to allow all the agents of an UAVs swarm to reach their targets with minimal interferences.

\citet{9108245} presents a recent survey discussing collision avoidance strategies for UAVs, which use different types of sensors.

Alternatively, it is possible to extract the data about objects' edges/salients from RGB images and use it to prevent collisions, either explicitly or implicitly, during the training of the agent's controller \cite{soria2020dense, he2019bidirectional, Zhang_2020_CVPR}.

In this section methods that alert of an up-coming collision, give direct control commands that prevent collisions, and collision safe multi-agent formation methods are presented. We see those approaches as complementary/safety tools for approaches like ours. These approaches mainly focus on where not to go, while approaches like ours focus on where an agent should go, out of a larger number of options.

\subsection{Deep learning for feature extraction and depth prediction}
Deep learning is a very active and developing area used for computer vision, useful for various tasks. Here, we briefly mention the most relevant deep learning approaches, directly utilized or relevant for this work.

DNNs allow high-quality feature extraction from natural images, and a sequence of increasingly improving performance, including AlexNet \cite{krizhevsky2012alexnet}, VGG \cite{simonyan2014very}, ResNet \cite{he2016resnet}, Inception-ResNet \cite{szegedy2017inception}, and DenseNet \cite{huang2018densely}. Training of those networks on a large-scale datasets can provide a meaningful features for any natural image input.

\cite{Dhamo_2019} takes a RGB image as an input, and predicts both depth data and foreground separation masks. This allows them to predict how a scene will look without the foreground objects, and therefore, for example, how only the walls of a room will look, without the current objects which exist in it.

The extracted features can be further used for a variety of tasks, including classification, regression, object detection, etc. In particular, when followed by upsampling decoder, image-to-image translation tasks can be solved efficiently \cite{badrinarayanan2015segnet,ronneberger2015unet}. An example of such image-to-image task is depth prediction from a RGB image input \cite{alhashim2019high}, and \cite{DBLP:journals/corr/abs-2011-14141} which follows and uses a transformer-based architecture block that divides the depth range into bins whose center value is estimated adaptively per image. More RGB image to depth map prediction methods can be seen at \cite{10.1007/978-3-030-20870-7_41} and \cite{7785097}.
\section{GoToNet Framework}
\label{sec:method}
To solve the problem of scene exposure and exploration using only an RGB camera, we propose to use a novel dataset generator. A generated data sample consists of RGB-D images, along with two (x,y) pixel coordinates for Goto and Lookat pixels. These two pixels represent flying directions, and are optimized to expose the largest amount of currently hidden areas.

\begin{figure*}
\centering
\includegraphics[width=\linewidth]{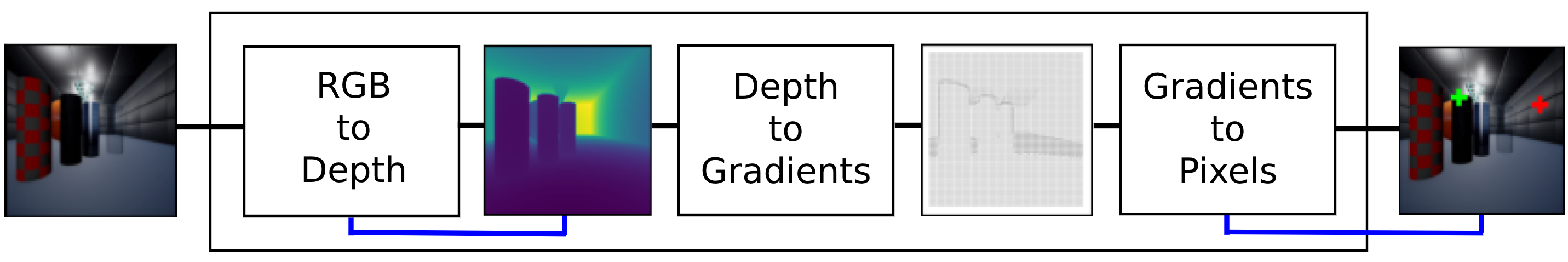}
\caption{Navigation model overview. The input of the system (on the left) is an RGB image. The RGB image is translated into a depth image by the DenseDepth model. Then, using the Sobel--Feldman operator, the gradient map is extracted from the predicted depth image. Finally, the gradient-to-pixel DNN predicts  Goto and Lookat pixels. The blue lines represent the gradient flow during training.}
\label{fig:system_overview}
\end{figure*}

Our model (visualized in Figure \ref{fig:system_overview}), described in \cref{sec:gotonet}, is trained to predict these coordinates in three steps:
\begin{enumerate}
    \item Predict a depth map from an RGB image.
    \item Generate a gradient map from the depth map.
    \item Predict the Goto and Lookat pixel coordinates from  the gradient map.
\end{enumerate}

The depth images are used to train the first part of the model, thus only RGB input is required in inference time. In case depth images are available during inference, ground truth depth can be used instead of the predicted depth map.

\subsection{Dataset generation}
\label{sec:dataset}
To create a dataset, we use the Microsoft AirSim simulator \cite{shah2017airsim}, which is built on the Unreal Engine \cite{unrealengine}, using a scene we built. To create a data sample, we place the drone in a 3D pose, capture images using the RGB-D camera, save both RGB and depth images, and follow the steps explained in Section \ref{gtg_data} to generate the navigation pixels. Different parts of our model are trained on different parts of the dataset but, as a whole, the dataset consists of pairs of RGB-D captured images, and Goto and Lookat pixel coordinates.

\subsubsection{RGB to depth dataset}
Our system is trained using both RGB and depth images, and uses only RGB images during inference. Therefore, we train one of our DNNs to predict depth images from RGB input, using images captured by an AirSim drone's RGB-D camera.

\subsection{Goto and Lookat pixels' inference dataset}
\label{gtg_data}
This dataset is used to train the navigation model. The input in this case is a gradient map, which is acquired during inference by applying the Sobel--Feldman operator \cite{sobel1968} to the depth map. The label is composed of the Goto and Lookat pixel coordinates, a total of four values. To create the Goto and Lookat pixels, the following steps take place.

\subsubsection{Collisions detection} 
To calculate the 3D collisions points, we use the depth image. First, ignoring the depth values, we take all the (x,y) coordinates of the image ($224\times 224=50,176$), back-project, and normalize them into 3D unit vectors, which all start at the camera's center of projection, and point according to the pixels' locations at the camera's projection plane. Second, we multiply all of those unit vectors by their corresponding depth values. After this, each depth image's pixel, has a corresponding 3D vector, which start at the camera's center of projection, and ends at the estimated 3D point of it's collision with an object which is in the scene. We do this in order to have set of 3D points which can be seen from this camera pose, as data for the up-coming scene painting phase.

\subsubsection{Scene painting} 
To mark seen areas, we paint the areas in the scene that are visible from the current position. To this end, we triangulate these 3D surfaces such that the vertices are the 3D collision points inferred from each pixel's depth value, and the triangles' vertices correspond to neighboring pixels in the depth image. To reduce False-Positive triangles -- Triangles which are likely to fill empty space, and therefore are not parts of any of the objects' surfaces, we then use the gradient magnitude heuristic to drop suspicious triangles. This is done if the gradient between two neighboring pixels, corresponding to a triangle's vertices, is large (top 4\% of all gradients). These 3D points likely lie on different objects, and thus should not be connected.
The remaining triangles, which we denoted as suspicious, and therefore are not considered as legitimate triangles in 3D space, are separated into their vertices, and are then added to scene and painted to denote already seen areas in the scene using a simple 3D point in space.

\subsubsection{Inferring the Goto and Lookat pixels}
\label{inferring_goto_lookat_pixels}
After the visible parts of the scene are painted by a covering layer of 3D triangles, we calculate the Goto and Lookat pixels. To do this, we check how the scene looks from a number of possible camera poses, and use the information regarding the amount and distribution of unseen areas, in terms of pixels representing those areas, to determine the navigation instructions. 
The camera poses are chosen by $K=k_x\times k_y$ pixels $\vb{p}_{i,j}$ with coordinates $(x_{i}, y_{j})$ for $1\leqslant i\leqslant k_x$, $1\leqslant j\leqslant k_y$, which are uniformly spread over the image (we use $k_x=k_y=3$, meaning $K=9$, but larger values should improve performance at the expense of additional computing demands). For each of these $K$ pixels, we calculate a pose for the drone by back-projecting each pixel into a normalized 3D directional vector, which is then multiplied by a constant 1 unit of the simulator's distance. The result is a vector which starts at the camera's optical center, and ending at the inferred camera pose. An illustration can be seen in Figure \ref{fig:data_creation}.

From each of these $K$ camera poses, a new RGB image, denoted $I_{i,j}$, is captured. For each of these $K$ new RGB images, we want to count the pixels that represent unseen areas, $N_{i,j}=\abs{\mathcal{U}_{i,j}}$, where $\mathcal{U}_{i,j}$ is the set of unseen pixels for image $I_{i,j}$.
For each of $K$ samples, we define a weight $w$ equal to a squared portion of unseen pixels in this pose out of all unseen pixels:
\begin{align}
    p_{i,j} &= \frac{N_{i,j}}{\sum_{i,j} N_{i,j}}\\
    w_{i,j} &= \frac{p^2_{i,j}}{\sum_{i,j} p^2_{i,j}} = \frac{N_{i,j}^2}{\sum_{i,j} N^2_{i,j}}
\end{align}
The Goto pixel coordinates ($\vb{g}$) are then calculated as a weighted average of pixels coordinates that can be transferred into possible future locations of the agent. The Lookat pixel coordinates  ($\vb{l}$) are calculated as a weighted average of the centers of masses of the unseen pixels; both averages use weight $w_{i,j}$:
\begin{align}
    \vb{g} &= \sum_{i,j} w_{i,j} \cdot \vb{p}_{i,j} \label{eq:goto}\\
    \vb{l} &= \sum_{i,j} w_{i,j} \cdot \vb{m}_{i,j} \label{eq:lookat}
\end{align}
where $\vb{m}_{i,j}$ is the center of mass of unseen pixels, defined as
\begin{equation}
\vb{m}_{i,j} = \frac{1}{N_{i,j}} \sum_{\vb{u} \in \mathcal{U}_{i,j}} \vb{u}
\end{equation}
Using instructions based on a weighted average, instead of choosing the direction with the maximal number of unseen pixels, allows us to reduce the minimal value of $K$ required for a high-quality direction choice, and makes the agent's physical behavior smoother during flights.

To determine $\mathcal{U}_{i,j}$, we calculate the Euclidean distance in the CIELAB color space ($\Delta E$) between the pixels' colors, and the color that marks a pixel as seen (red). If a pixel's distance is smaller than a chosen threshold, it is declared as seen:
\begin{equation}
\mathcal{U}_{i,j} = \qty{\vb{p} \mid \vb{p} \in I_{i,j}, \: \alpha < D_c(\vb{p})},
\end{equation}
where $D_c$ is a color difference function that compares the color of the pixel $\vb{p}$ with the color used for seen pixels (red in Figure \ref{fig:data_creation}); $\alpha$ is a threshold. We used $\alpha=65$.
Usage of the threshold allows us to compensate for the simulator's visual effects, such as fog, which may cause a surface to appear with minor color changes. 

In order to handle the non-exposing case, where there is no camera pose, out of the $K=k_x\times k_y$ camera poses, that exposes a sufficient amount of unseen pixels, The Goto and Lookat pixels are chosen using the following method:

To detect this case, we iterate over all of the $K=k_x\times k_y$ camera poses. For each camera pose we sum the number of pixels which represent areas unseen from the original camera pose. If not one of the camera poses' unseen pixels count, divided by the total number of the image's pixels, is equal or higher than a chosen percentage threshold, it is declared as a non-exposing case. (We set the threshold to $1.5\%$)

If the non-exposing case was detected, both the Goto and Lookat pixels are set to be the pixel which holds the maximum depth value.
This is done in order to both make the flight continuous, and make the agent turn to look at hopefully the largest volume of space possible from it's position, in hope it will allow for a better futuristic unseen areas exposure. We noticed this case happens mostly when the agent faces a wall directly, and therefore the pixel which holds the deepest depth value also represents a way out of looking only at that wall.

Notice: Our method paints the scene in order to mark areas visible from the original camera pose. To mark those visible areas as seen, we also considered using 3D features detection methods, such as SIFT \cite{790410}, instead. This means that features detected from the original camera pose, and are visible from any of the $K=k_x\times k_y$ camera poses, will be considered as already seen areas. While features not visible from the original camera pose, but are visible from any of the $K=k_x\times k_y$ camera poses, will be considered as unseen areas. But, we came to the conclusion using 3D features, instead of painting the scene, is less reliable. This is because, for example, the case where there are 2 surfaces, where 1 of them is hiding the other, and both of them do not contain any features. If the camera moves to a location which does expose the hidden surface, but can not detect features laying on that hidden surface, it will not be able to conclude that this location actually does expose hidden surfaces.

\subsection{GoToNet -- Flight instructions framework}
\label{sec:gotonet}

The proposed framework comprises of three parts: depth prediction, gradient map extraction, and navigation. The first and the last parts are DNNs, while the second part is a classical computer vision algorithm.

\subsubsection{Depth map prediction}
\begin{figure}
\centering
\includegraphics[width=\columnwidth]{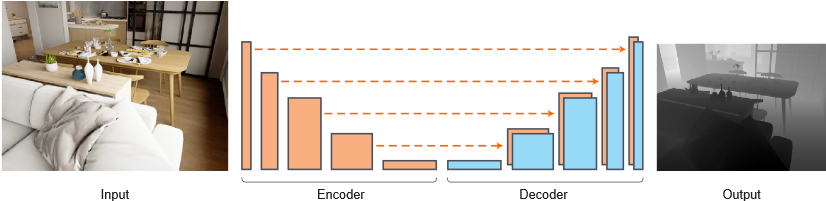}
\caption{The DenseDepth RGB-to-depth model architecture (figure by \citet{alhashim2019high})}
\label{fig:densedepthmodelarchitecture}
\end{figure}

\begin{figure}
\includegraphics[width=\columnwidth]{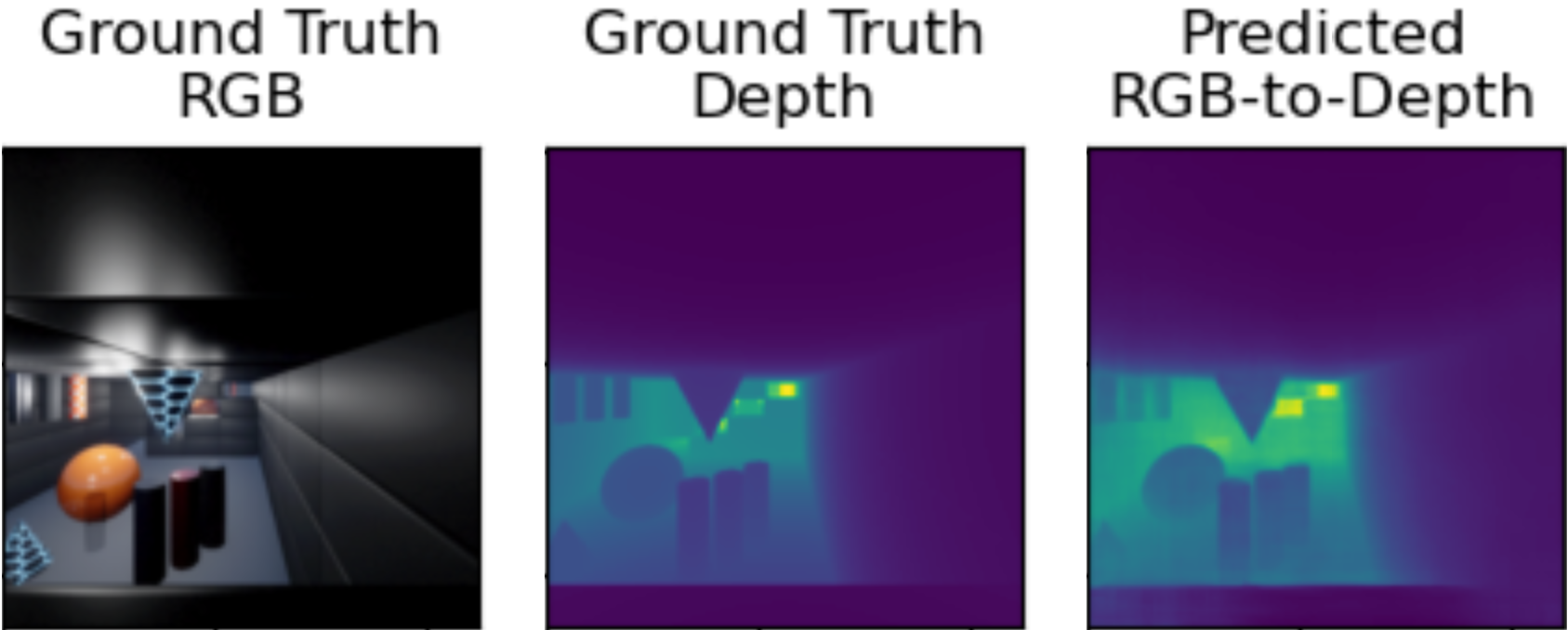}
\caption{An example of an RGB-to-depth prediction, using our test set. }
\label{fig:rgb_to_depth}
\end{figure}

We train the DenseDepth model \cite{alhashim2019high}  using the RGB-D part of our dataset to predict the depth from RGB input. DenseDepth is a UNet-like model, where the encoder is a DenseNet \cite{huang2018densely} and the decoder is made of a series of up-sampling layers. The model's architecture is visualized in Figure \ref{fig:densedepthmodelarchitecture} and the output example is shown in Figure \ref{fig:rgb_to_depth}.

\subsubsection{Gradient map extraction}
\begin{figure}
\centering
\includegraphics[width=\columnwidth]{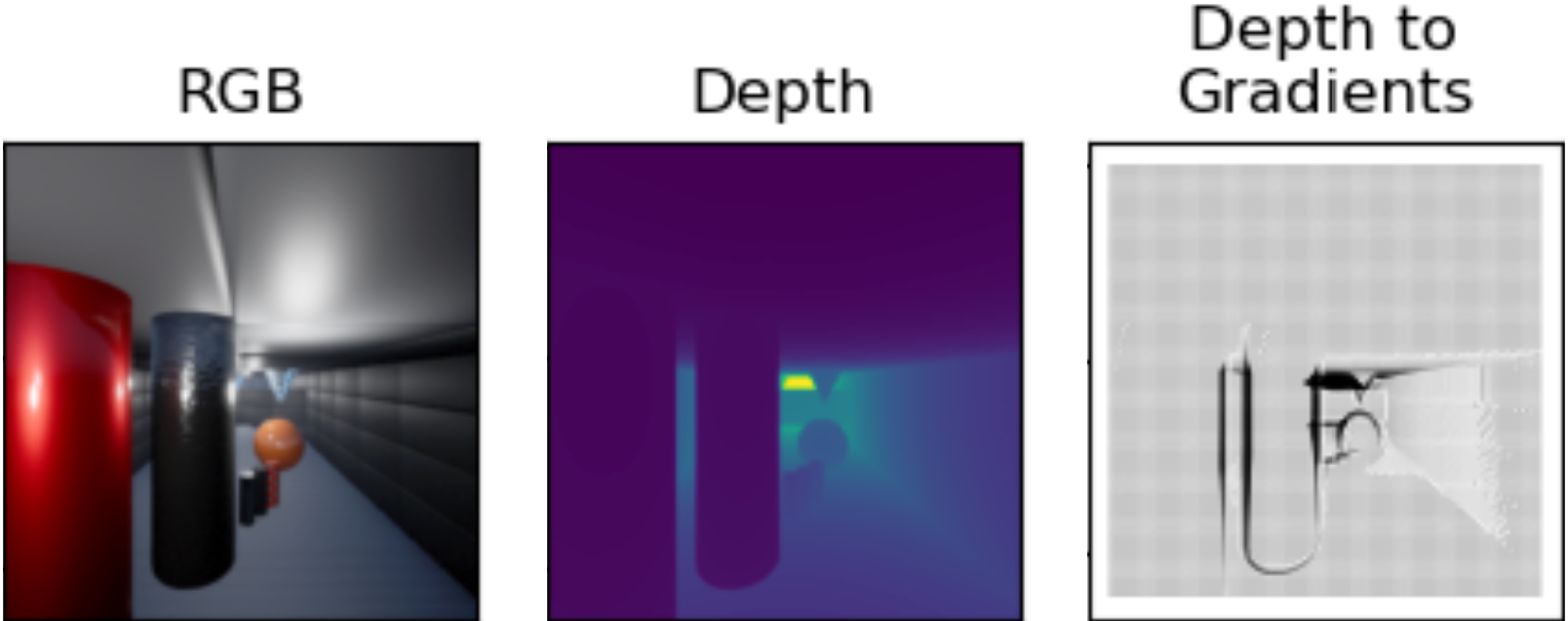}
\caption{An example of the creation of a gradients map. \textbf{Left}: The RGB image. \textbf{Middle}: The input depth image. \textbf{Right}: The output gradients map. We create gradient maps using the outputs of the RGB-to-depth model as inputs.}
\label{fig:gradients_label}
\end{figure}

To transform a depth image, either generated in a previous stage or provided by a depth camera, into a gradient map, we utilize the Sobel--Feldman operator \cite{sobel1968}.
A gradient map of a depth image represents the change of depth between neighboring pixels. For example, for the $x$ direction, the gradient map is the approximation of $\nicefrac{\partial d}{\partial x}$ where $d$ is the depth function. A gradient map is useful since sharp gradients may indicate a large open space hidden by an object, while a raw depth image does not directly provide information about depth changes between neighboring pixels. Implicitly learning relevant depth information directly from the RGB input, with only navigation instructions as supervision, is a significantly harder task and is prone to overfitting.
An example of the usage of this algorithm can be seen in Figure \ref{fig:gradients_label}. 

\subsubsection{Navigation instruction generation}

\begin{figure}[htp]
\centering
\includegraphics[width=\columnwidth,height=\textheight,keepaspectratio]{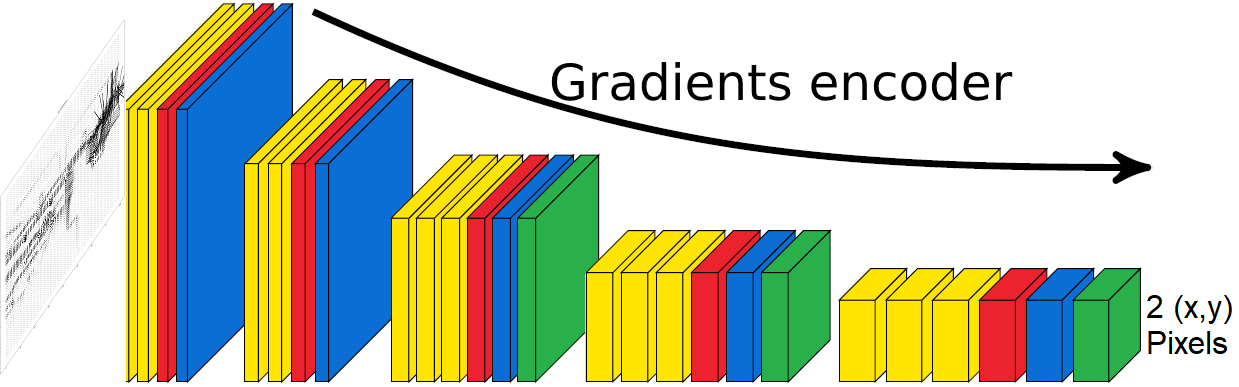}
\caption{The architecture of the model which maps gradients to Goto and Lookat pixels (the figure is based on a figure taken from \cite{10.1007/978-3-319-54181-5_14}).}
\label{fig:gradients_to_pixels_model}
\end{figure}

As explained earlier, our framework uses Goto, and Lookat pixels as navigation instructions for the agent.
We use a custom modification of the VGG16 network \cite{simonyan2014very}, implemented in a fully convolutional network manner \cite{long2015fully}. We take 5 of the pretrained VGG16 netowork's convolutional layers, end each of them with batch-normalization, and a ReLU activation layers, making them the so called \textit{CBR layers} (convolution -- batch normalization -- ReLU). After those 5 pretrained CBR layers, we add our own 2 not-pretrained additional CBR layers, making the network contain a total of 7 CBR layers. The final output contains four values, which represent the Goto, and Lookat pixels. This network (visualized in Figure \ref{fig:gradients_to_pixels_model}) is trained on 2-channel images (where the channels are for the gradients maps, with respect to $x$ and $y$). The network outputs two pairs of coordinates, and is trained to minimize the MSE loss between the output and the Goto, and Lookat pixels' coordinates, calculated as described in Section \ref{sec:dataset}.

\begin{figure}
\centering
\includegraphics[width=\columnwidth]{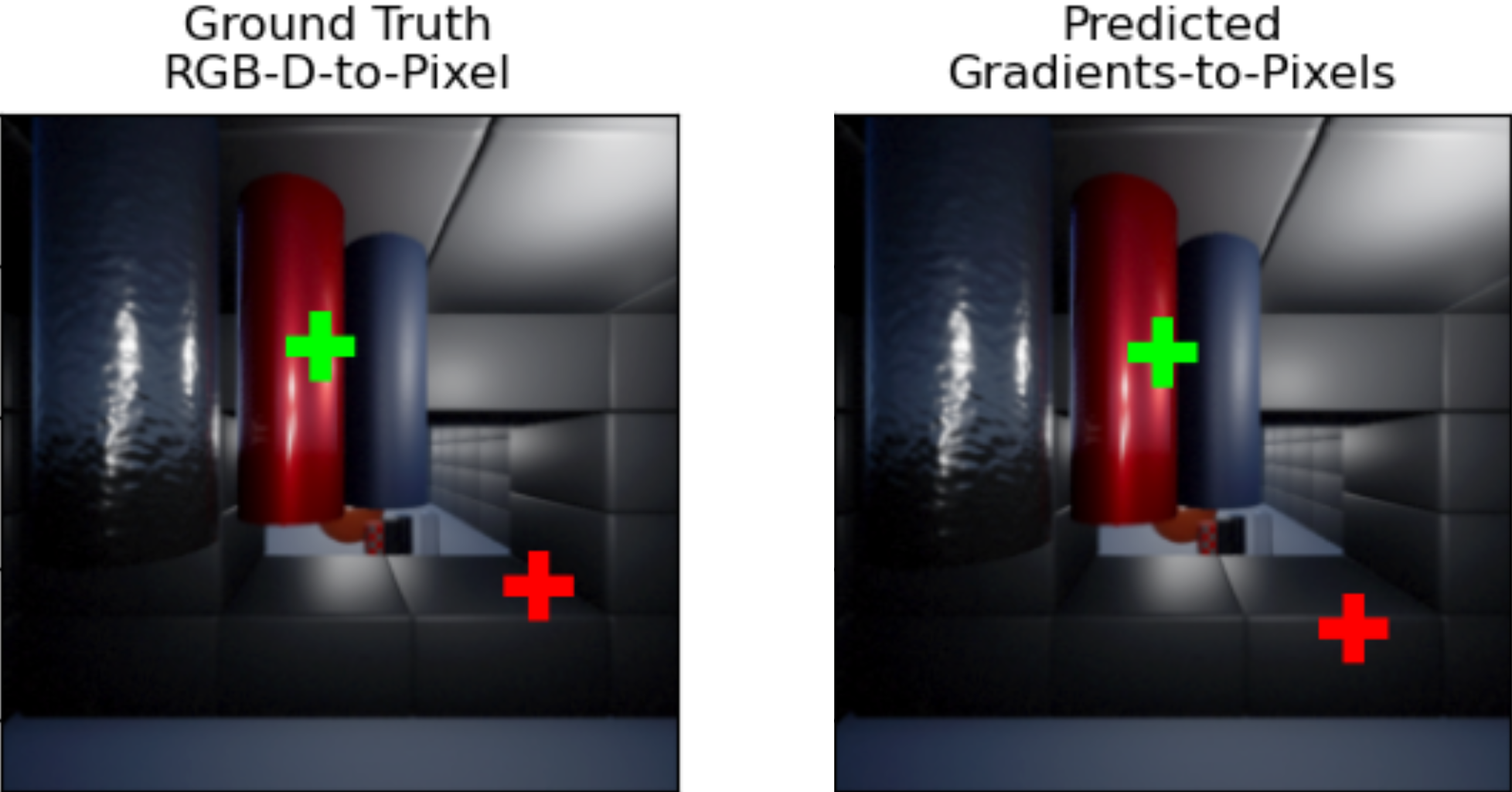}
\caption{\textbf{Left:} An example of a data sample, which was calculated using an RGB-D camera's output in the simulator, and which shows the navigation  Goto (red) and Lookat  (green) pixels, determining movement and rotation directions, correspondingly. \textbf{Right:} The navigation pixels predicted by our model using only an RGB image as an input.}
\label{fig:simulator_vs_model}
\end{figure}
\begin{figure}
\centering
\includegraphics[width=\columnwidth]{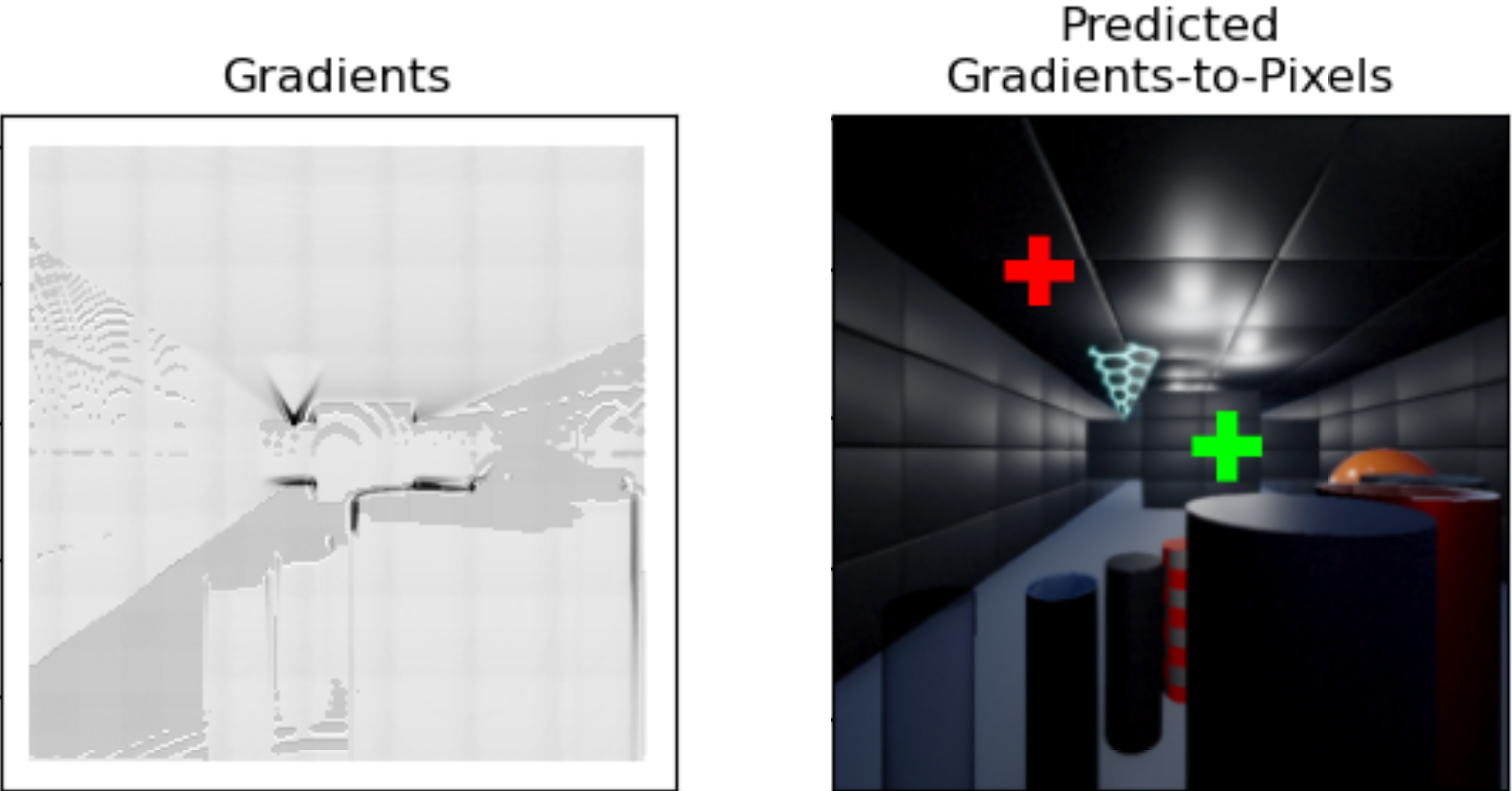}
\caption{An example of the gradients map to Goto Pixel Model's prediction. The red marker is the Goto pixel, and the Green marker is the Lookat pixel.}
\label{fig:gradients_to_pixels_example}
\end{figure}

During the inference stage (shown in Figures \ref{fig:simulator_vs_model} and \ref{fig:gradients_to_pixels_example}), the agent backprojects the Goto and Lookat pixels into 3D vectors. Usage of these vectors is illustrated in Figure \ref{fig:goto_lookat_movement}. The movement is determined by Goto vector: the agent moves in the direction of the Goto vector by a single length unit. The Lookat vector is used for rotation. We restrict the agent rotation to the horizontal plane, i.e. only yaw rotation is performed. Thus, we project the Lookat vector onto the horizontal plane and determine the angular difference between the current principal axis and the projected Lookat vector. The agent is then rotated by the resulting angle, which is restricted to the camera's FOV. Rotation and movement are independent and can be performed simultaneously.

\section{Experiments}
\label{sec:experiments}
To evaluate the performance of the proposed algorithm, we both compare how well is our sparse data creation process, described in \cref{inferring_goto_lookat_pixels}, compared to a more dense case, and also we run the model in our environment and measured how well it is able to explore the scene at the basic case of 2D exploration, and the more advanced 3D exploration case.

We compare our approach to Learn-to-Score \cite{hepp2018learntoscore} -- a voxel-based, next-best-view, DNN-based method. Our method works in more challenging settings, using RGB input instead of voxels, which allows simpler transferring into real-world settings. In addition, having continuous action space and performing translation and rotation in a single step to any point in the camera's FOVs allows us to make better micro-decisions, with more accurate pose changes.

In this section we explain the experiments' settings, including some of the hyperparameters used (Section \ref{experimental_settings}), the ways we evaluate both methods (Section \ref{evaluations}), and finally the comparisons between the two methods (Section \ref{comparison}).

\subsection{Experimental settings}
\label{experimental_settings}
To test our approach, we use a Microsoft AirSim's drone on a custom-made Unreal Engine indoor scene, which includes many rooms, objects, and types of ways to move between them.

To check how well is our sparse data creation process, Section \ref{pixels_inferring_evaluation} describes how we compare it to a more dense case, where the inferred pixels are inferred using a higher number of camera poses.

To check how well our framework works as a whole, we conduct 1000 test runs, placing the drone in a random starting pose, laying a target in some random 2D positions (Section \ref{2d_evaluation}), or 3D positions (Section \ref{3d_evaluation}), and letting the drone explore the scene. The run terminates if (1) the maximum number (500) of steps is reached, (2) the amount of novel voxels seen by drone is less than the threshold (30) for 20 consecutive steps, or (3) the drone collides with any object's bounding box. Note that during the  dataset creation a collision is defined as actually coming into contact with the object's boundaries, but during testing  it is defined as crossing the object's boundary box. This makes the conditions stricter than what the framework is trained to do. 

For our method we capture an RGB image and use the image as an input to our navigation model to predict the Goto and Lookat pixels. We then use the Goto pixel to fly along for 1.0 AirSim distance unit and use the Lookat pixel to rotate accordingly.

For Learn-to-Score we use a depth camera to check all possible next steps and choose the step with the highest score, moving 1.0 AirSim distance unit, or rotating around the yaw axis by 25 or 180 degrees.

\subsection{Evaluations}
\label{evaluations}

We evaluate the performance of our approach by measuring various scene exposure metrics in 2D and 3D environments. In addition we perform an ablation study to justify the choice of the hyperparameter $k$.

\subsubsection{Pixels Inferring Evaluation}
\label{pixels_inferring_evaluation}
Here we evaluate the performance of the proposed sparse data creation process, described in \cref{inferring_goto_lookat_pixels}, compared to a denser case. We first randomly pick 1183 camera poses, and for each of them, we use our method to determine the Goto and Lookat pixels. For comparison with the $k_x=k_y=3$ case used in the paper, we try to increase the number of poses to $k_x=k_y=9$, and instead of using our process for sparse data, we simply choose the Goto and Lookat pixels corresponding to the camera pose which possesses the maximum amount pixels that represent unseen areas. The optimal calculation, utilizing all $K=224\times 224$ possible camera poses, is computationally infeasible. The difference between the methods is quantified in distance between the Goto and Lookat pixels, in units of screen diagonal length.
We conclude that although given a ninefold computational advantage over the sparse method, our averaging method indeed provides good results, that are close to an optimal case's results.

The results can be seen in Table \ref{table:evaluation_gt}.

\begin{table}
\centering
\begin{tabular}{lcc}
\toprule
Measurement&Distance between $k=9$ and $k=3$\\
\midrule
 Goto pixel  & 0.208\eb{0.104}\\
 Lookat pixel  & 0.021\eb{0.0194}\\
\bottomrule
\end{tabular}
\caption{A comparison between proposed inference method using $k_x=k_y=3$ and a case which selects from $k_x=k_y=9$ camera poses the one with the maximum number of pixels that expose unseen areas, in units of screen diagonal length, in the format of mean\eb{std}.}
\label{table:evaluation_gt}
\end{table}

\subsubsection{2D Evaluation}
\label{2d_evaluation}
\begin{figure}
\centering
\includegraphics[width=\columnwidth]{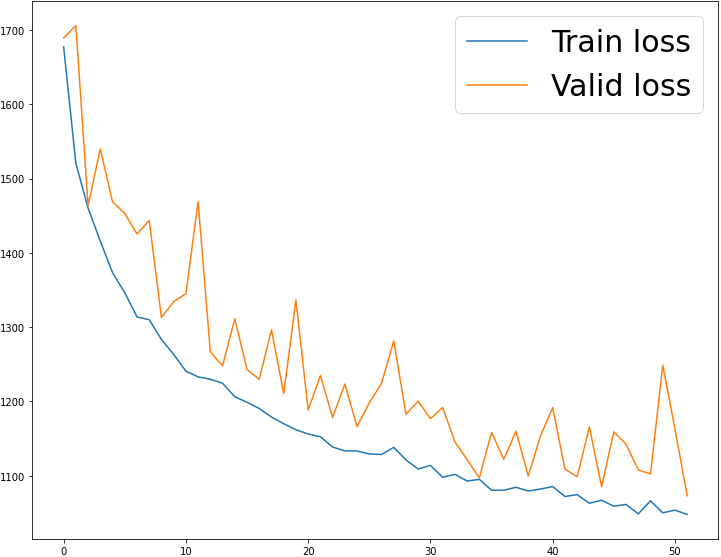}
\caption{The training results of the Gradients-to-pixel DNN model's MSE loss vs.\ epoch number, for the 2D case. The numbers here are higher compared to the 3D case, since the ground-truth pixels labels lay on an 1D line (Go left-straight-right, not up/down), while the model still can predict 2D pixels.}
\label{fig:fit_results_2d}
\end{figure}

We start from a simpler, limited scenario in which the drone is limited to move only on a 2D horizontal plane. here we give only partial explanations compared to the 3D case. 
As opposed to the full 3D dataset, in this case we limit ourselves to $K=1\times k_y$ camera poses (instead of $K=k_x\times k_y$). 
The training set consists of 37,172 data samples. For training, the data was split into training (85\%) and validation (15\%) datasets.
We used learning rate of  0.0005, momentum: 0.9, and weight decay of 0.0005. Training results can be seen in \cref{fig:fit_results_2d}.

The GoToNet model 2D evaluations results can be seen in \cref{table:evaluation_2d}.

\begin{table}
\centering
\begin{tabular}{lcc}
\toprule
Measurement&Ours\\
\midrule
 New voxels per camera pose & 918.61\eb{368.04}\\
 Minimum distance to target & 327.46\eb{244.47}\\
 Percentage of surface voxels seen & 4.92\%\eb{5.69\%}\\
\bottomrule
\end{tabular}
\caption{2D evaluation of our approach in the format of mean\eb{std}.}
\label{table:evaluation_2d}
\end{table}
\subsubsection{3D Evaluation}
\label{3d_evaluation}
After evaluation the method on it's own in the 2D case, here we evaluate it in the 3D case, and also compare it to the baseline method. We train both methods by creating a dataset for each method in the Unreal Engine simulator, while using the same scene we designed for both. Both methods were trained under as close as possible conditions to each other. All cameras have FOV angles of 90 degrees along both axes, and resolutions of $ 224 \times 224 $ pixels.

\begin{figure}
\centering
\includegraphics[width=\columnwidth]{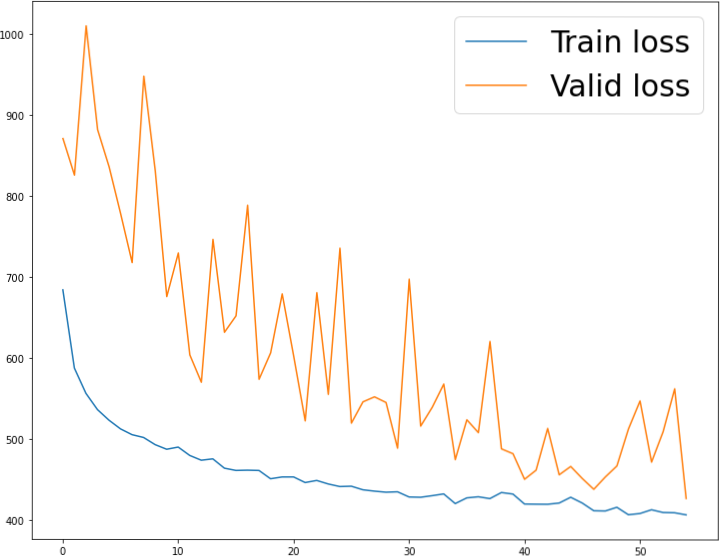}
\caption{The training results of the Gradients-to-pixel DNN model's MSELoss Vs. epoch number, for the 3D case.}
\label{fig:fit_results}
\end{figure}

Learn-to-Score is trained on a dataset of 256,200 data samples, without using any augmentations, as described in the original paper. 
Since for GoToNet data generation we add additional objects during the painting phase, which slows down the rendering, we use a smaller dataset for our model. The training set consist of 48,400 data samples. For training, the data is split into training (85\%), and validation (15\%) datasets. We use the following hyper-parameters: learning rate 0.0005, momentum 0.9, weight decay 0.0005. Training results can be seen in \cref{fig:fit_results}.

The resolution of both RGB and depth images during train and test is $ 224 \times 224 $ pixels. The choice of resolution is a critical one: while the lower resolution enables faster inference, a larger one should improve the model's decision making. For example, with higher resolution input, the model will be able to detect sharp depth changes in the gradient map.
During training, results on the validation dataset indicate that the root mean squared error for the Goto and Lookat pixel coordinates is 29.23 pixels on a $ 224 \times 224 $ image, i.e.,  \textasciitilde13\% of the image's horizontal/vertical size, or \textasciitilde9\% of the image's diagonal size.

During the experiments, the starting poses of the test flights, which the flight paths depend on, are randomly picked, and, therefore, are unlikely to appear in any dataset.

\begin{figure}[htp]
\centering
\includegraphics[width=\columnwidth,]{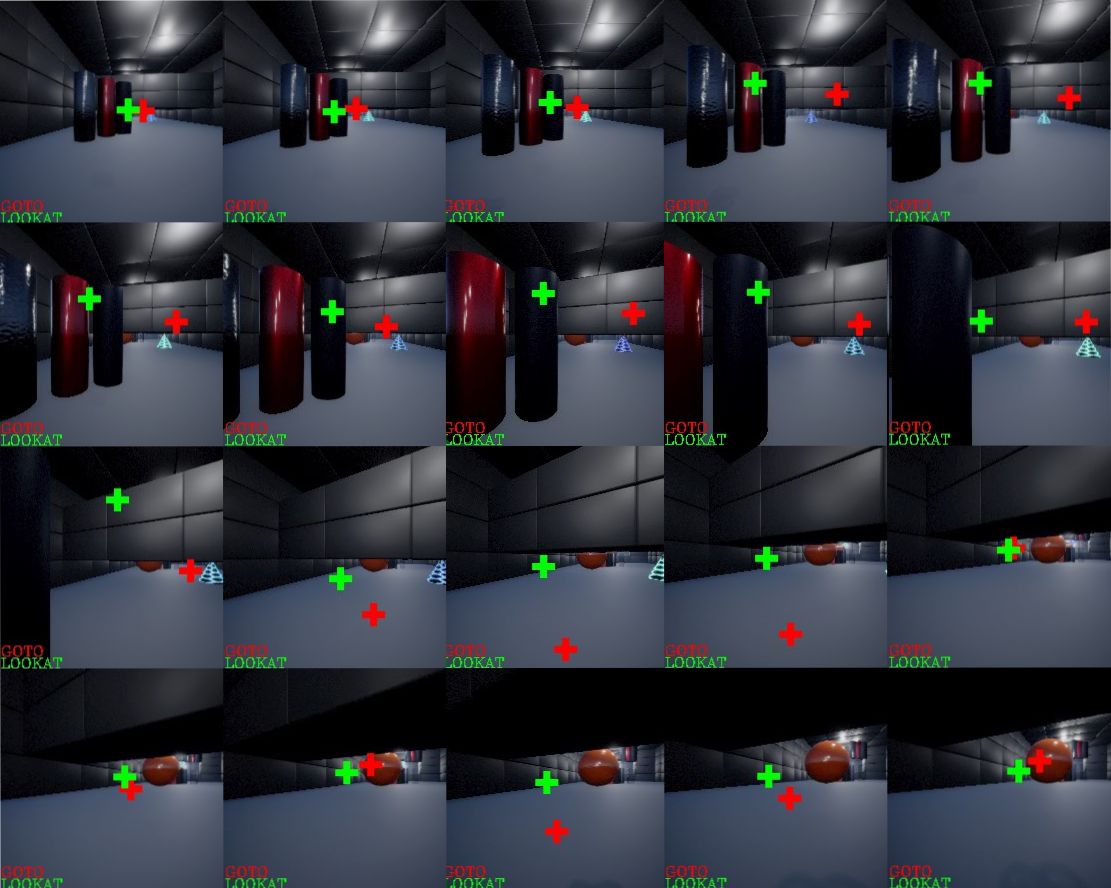}
\caption{A test flight at different timesteps (time progress: left to right, then top to bottom). This is not one of the 1000 flight test described in Subsection \ref{experimental_settings}.}
\label{fig:Flight_1}
\end{figure}

For both neural networks, we use partially pretrained networks, as described in \cref{sec:gotonet}. The RGB-to-depth model uses ImageNet-pretrained DenseNet-161 \cite{huang2018densely} as the encoder and the gradients-to-pixels is a modified VGG \cite{simonyan2014very}. Then, as a complete model, we also pretrain the gradients-to-pixels model using intermediate versions of the full dataset, built from earlier versions of the scene and containing less samples. Later, the whole model is fine-tuned on the final version of the dataset.
During training we use data augmentations, which included random horizontal flips and a random cropping of at least 90\% of the original image (a test flight is illustrated in Figure \ref{fig:Flight_1}).

\subsection{Comparison}
\label{comparison}
We compare the methods using three metrics, described below.

To evaluate how much of the scene is exposed, we divide the scene into voxels. For each voxel, we store a boolean variable $S_v$, which determines whether it was already seen or not. At each step, we perform ray tracing to determine which voxels are seen and update the boolean variables accordingly. We also count the number of voxels for which $S_v$ was updated, i.e., the number of voxels seen for the first time at this time step.   

We also evaluate the average minimum distance between randomly placed targets and the agent's position along the path. Each test has one target and, therefore, one closest-to-target agent's position along the path. In this scenario, in both methods, we test a case where the drone, other than exploring, should also be able to find, or get close to something, while only having a camera (RGB, or depth) as its data source. The reason we chose to evaluate using this metric, is to check how well an agent advances in the scene between different locations and increases its chances of contacting a target without previous knowledge about its location.

Finally, we evaluate the average percentage of voxels each method exposes, out of all the scene's occupied surface voxels. For this, we determine which voxels intersect with bounding poxes of any of the objects in scene. The voxel is considered a surface voxel if it has at least one unoccupied  neighboring voxel. Of these voxels, we count the voxels that were seen by our agent during the episode, i.e., for which $S_v=1$ at the end of the episode. 
The average results are presented in \cref{table:evaluations}.
These three metrics represent how well a scene is exposed, in terms of both per-image exposure and total exposure, and how easily a randomly placed target can be discovered, without previous knowledge about where it may be.

\begin{table}
\resizebox{\columnwidth}{!}{\begin{tabular}{lcc}
\toprule
Measurement&Learn-to-Score \cite{hepp2018learntoscore} &Ours\\
\midrule
 New voxels per camera pose & 113.15\eb{60.42}&\textbf{961.93\eb{208.65}}\\
 Minimum distance to target & 340.59\eb{244.99}&\textbf{303.79\eb{239.98}}\\
 Percentage of surface voxels seen & 0.89\%\eb{0.80\%}&\textbf{8.89\%\eb{7.60\%}}\\
\bottomrule
\end{tabular}}
\caption{Comparison between Learn-to-Score and our approach in the format of mean\eb{std}. }
\label{table:evaluations}
\end{table}

\subsubsection{Metric design}
In addition to voxel-based metrics, we attempt to implement pixel-based metrics, using scene paining similar to dataset generation process. Unfortunately, painting the scene during evaluation is prohibitively expensive in terms of required resources for rendering and runtime, even though pixel-based metric would allow more accurate and realistic evaluation of our method.


\subsubsection{Runtime comparison}

\begin{table}[htbp]
\resizebox{\columnwidth}{!}{\begin{tabular}{lcc}
\toprule
Measurement&Learn-to-Score \cite{hepp2018learntoscore} &Ours\\
\midrule
Total runtime (GPU-minutes) & 36,389 &\textbf{7,809}\\
Total number of steps & 70,554 &\textbf{82,492}\\
Mean step time on RTX 2080 Max-Q (seconds) & 31.58 &\textbf{5.67}\\
\bottomrule
\end{tabular}}
\caption{Comparison of runtime between Learn-to-Score and our approach. }
\label{tbl:time}
\end{table}

Our method is run entirely on a laptop computer with Nvidia RTX 2080 Max-Q GPU. Due to large runtime of Learn-to-Score, we parallelize it on additional PC with RTX 2080 Ti GPU. The runtime data is summarized in \cref{tbl:time}. In addition to better performance, our method is almost an order of magnitude faster, which makes it more feasible to be used in a real-life setup. 
\section{Conclusion}
\label{sec:conclusion}
This work proposes a novel datasets generator and a DNN-based prediction model framework for fast scene exposure and exploration that provides movement (Goto pixel) and rotation (Lookat pixel) instructions for aerial vehicles, designed to maximize the coverage of previously unseen areas of the scenes.
The way the system works allows the user to use a single RGB image, and make optimized decisions, as opposed to many existing SLAM-based methods that try to map the current location, and only then make movement decisions.

The data creation process in the simulator steps: 1. 3D Painting of the visible areas. 2. Evaluating a number of possible next steps/camera poses for scene exposure qualities. 3. Infer the navigation recommendation by weight averaging the pixels counts from the camera poses. This allows for great flexibility which is not possible in the real world.

After training, our model requires only RGB data. As a result, it obviates the need for extra sensors, and their limitations. This allows the usage of smaller carriers, which other methods do not allow.

During testing, we compared our sparse case of 3 X 3 camera poses to a 9 X 9 camera poses case, in which we inferred the instructions by choosing the pixels which go to the camera pose which contains the maximal number of pixels representing unseen areas. The results indicated our weighted average approach came close to the much denser case.

We compared our method in 3D test flights to a voxel-based next-best-view scene exposure method. In the new voxels per camera pose metric, our method exposes 850\% more voxels per camera pose. In the minimum distance to target metric, our method brings the drone 10.5\% closer to the target. In the percentage of the scene's surface voxels seen metric, our method exposes 998.8\% more voxels. In the compute time metric, our method calculates a command using 17.95\% of the time it takes for the baseline approach. These results indicate that our approach significantly outperforms it.

Transferring models from simulations to the real-world settings is a separate problem, which is extensively studied lately \cite{hofer2020perspectives,8861136,farahani2021reviewda} and is outside of the scope of this paper. However, given the general advancement of the sim-to-real approaches to computer vision and high realism of the simulator, we believe that transferring the approach to real-life settings is indeed feasible. Yet, the main limitation for testing this hypothesis is not adapting the trained network to the real life, but rather creating a polygon suitable for the testing of the resulting agent in various settings, as well as choosing and measuring evaluation metrics.

\section{Future work}
\label{sec:future_work}
The major possibility that we think will increase productivity the most is developing our system to have cooperative multi-agent capabilities. The problem is: How can agents share what they sense, and how do they divide their next action in such a way that maximize the overall exposure per time-step ? Try to develop our system in a more end-to-end approach, that may decrease the size of the overall model, and therefore the runtime as well. Other possibilities mainly include optimizations: Increase the number of camera poses used to decide the next-best-step from currently only 3 X 3 camera poses. Increase the accuracy of our seen-areas painting algorithm, so it better minimizes the number of 3D edges that stretch inside empty spaces, instead of upon the scene's objects. Evaluating the system with future state-of-the-art deep neural networks, to evaluate how well the quality of the system increases.

\backmatter

\section*{Declarations}

\subsection*{Author Contributions}
All authors contributed to the study conception and design. Material preparation, data collection and analysis were performed by Tom Avrech. The first draft of the manuscript was written by Tom Avrech, edited by Evgenii Zheltonozhskii, Chaim Baskin and Ehud Rivlin. All authors commented on previous versions of the manuscript. All authors read and approved the final manuscript.


\bibliography{gotonet}


\end{document}